# Neuromorphic Online Clustering and Its Application to Spike Sorting

J. E. Smith

University of Wisconsin-Madison (Emeritus)

June 14, 2025

## Abstract

Active dendrites are the basis for biologically plausible neural networks possessing many desirable features of the biological brain including flexibility, dynamic adaptability, and energy efficiency. A formulation for active dendrites using the notational language of conventional machine learning is put forward as an alternative to a spiking neuron formulation. Based on this formulation, neuromorphic dendrites are developed as basic neural building blocks capable of dynamic online clustering.

Features and capabilities of neuromorphic dendrites are demonstrated via a benchmark drawn from experimental neuroscience: spike sorting. Spike sorting takes inputs from electrical probes implanted in neural tissue, detects voltage spikes (action potentials) emitted by neurons, and attempts to sort the spikes according to the neuron that emitted them. Many spike sorting methods form clusters based on the shapes of action potential waveforms, under the assumption that spikes emitted by a given neuron have similar shapes and will therefore map to the same cluster.

Using a stream of synthetic spike shapes, the accuracy of the proposed dendrite is compared with the more compute-intensive, offline *k*-means clustering approach. Overall, the dendrite outperforms *k*-means and has the advantage of requiring only a single pass through the input stream, learning as it goes. The capabilities of the neuromorphic dendrite are demonstrated for a number of scenarios including dynamic changes in the input stream, differing neuron spike rates, and varying neuron counts.

## 1. Introduction

The primary goal of neuromorphic computer architecture is to discover the principles that underly neural computation in the brain and then apply the same principles to the development of advanced human-engineered computing devices that have brain-like capabilities and efficiencies.

Discovering the way biological computation is done has been a challenge in neuroscience for over a century, and applying biological principles to engineered computing devices goes back at least 80 years. One of the first proposed neuromorphic computing elements, the one that inspired artificial neural networks, is a single point model due to McCulloch and Pitts in 1943 (Figure 1). It was an attempt at modeling the basic components of a biological neurons: dendrites, synapses, neuron bodies, and axons. Although it oversimplifies the operation of a neuron as we know it today, it established a neuromorphic approach to the problem.

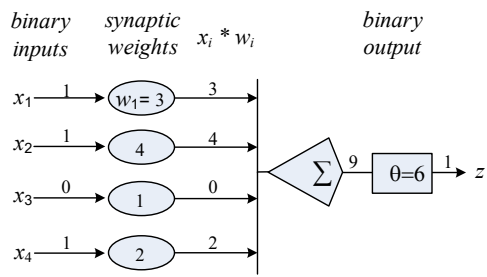

Figure 1. McCulloch and Pitts point neuron model (1943). Input spikes are represented as 1's. During inference, the dot product of the inputs and synaptic weights is first computed. Then, if the result exceeds a threshold value θ an output of 1 is produced; otherwise the output is 0.

A neuromorphic architecture as envisioned here is based on well-defined layers of abstraction. Working from the bottom up, the first four layers are: dendritic segments, dendrites, neurons, and columns. This paper focuses only on the bottom two, based on the proposition that dendrites are the fundamental building blocks for neural computation.

When reduced to a logical computing device, a biological dendrite becomes a *neuromorphic dendrite* ($n$D) that: 1) streams feature vectors as inputs, 2) learns similarities in an online manner, 3) groups similar feature vectors into clusters, and 4) outputs a stream of their associated cluster identifiers.

The primary objective of this paper is to demonstrate the features of an $n$D through a series of simulations. The simulation benchmark is drawn from spike sorting, an actively researched application in experimental neuroscience. Spike sorting is used in experimental studies directed at understanding the way neuronal aggregates interoperate. The experimental approach involves implanting an array of electrodes into the brain of a laboratory animal or growing cultured neurons over a substrate containing an embedded array of electrodes. Then as neurons under study are stimulated, either by the electrodes or some external source, neuron interoperation takes place, and emitted electrical signals collected at electrodes are analyzed in order to associate voltage spikes (action potentials) with individual neurons. That is, streams of spikes detected at the electrodes are sorted according to the neuron that emitted them. The objective of experimental scientists is to correlate output behavior with input stimulation, thereby learning something about neural functioning.

As part of many spike sorting algorithms, the shape of an action potential's waveform is a factor in distinguishing the emitting neuron. To do this, streamed-in waveform shapes are placed into clusters according to similarity, based on the underlying assumption that the shapes belonging to a given cluster are emitted by the same neuron. Furthermore, the ability to sort spikes in real time is an advantage for some experimental applications and essential for others such as brain computer interfaces (BCIs). This makes clustering waveform shapes an ideal application for demonstrating the capabilities of an $n$D.

## 2. Active Dendrites

Active dendrites [3] are part of an overall framework for biologically plausible neural networks that continually learn and adapt as they operate [9].

Biological excitatory neurons have the structure illustrated in Figure 2. The neuron has multiple dendrites, each divided into segments. Each segment has 10s of synapses, and the total number of synapses is on the order of thousands per neuron. Roughly speaking, synapses in an active dendrite model behave as they do in the classic point neuron model. However, in order to affect the neuron's body potential and trigger an output spike, simultaneous input spikes must occur at multiple synapses in close physical proximity, i.e., belonging to the same segment. Collectively, these spikes generate a dendritic pulse that travels along the dendrite to the neuron body. (Localized pulse generation is what makes the dendrites "active"). Consequently, by adjusting synaptic weights, synapses situated in a segment learn to detect a set of similar patterns, and when one of the patterns appears, a dendritic pulse is generated. Because a set of similar patterns forms a cluster, a dendritic segment performs a clustering function, and a dendrite consisting of multiple segments divides the input stream into multiple clusters.

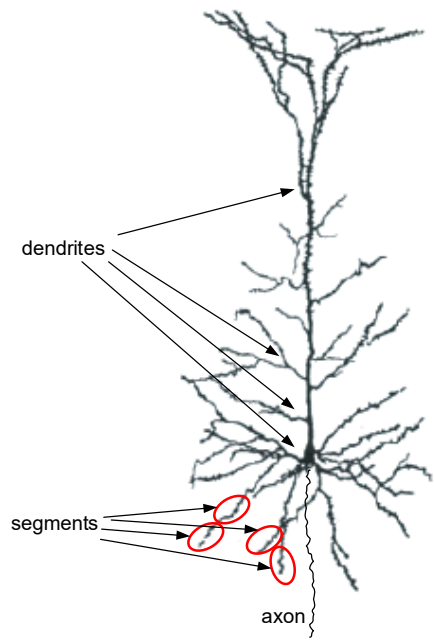

**Figure 2. An excitatory neuron consists of multiple dendrites, each of which is divided into multiple segments. In a multi-point neuron model employing active dendrites, multiple spikes converging on synapses within the same segment generate a pulse that propagates to the neuron body. This pulse (possibly in conjunction with other inputs) causes the neuron to produce an output spike on its axon. In contrast, spikes that are spread out over multiple segments are not sufficient for generating the necessary dendritic pulse.**

In a complete active dendrite model, there are two types of synapses: proximal and distal. As espoused by Hawkins and Ahmad [3], a proximal synapse acts as an enable signal of sorts. A proximal synapse is close to the neuron body, or attaches directly to the neuron body, so that a single spike at a proximal synapse is sufficient to trigger a neuron spike. A simultaneous pulse caused by distal synapses modulates the strength and time of the neuron's output spike. Because this paper goes no further up the neural hierarchy than a single dendrite, the dendrite in this paper is effectively always enabled, so there are no proximal synapses.

As noted earlier, this paper focuses only on the bottom two abstraction layers of a larger neuromorphic architecture. Other documents written by the author describe higher levels. In brief, at the third architecture layer multiple dendrites are combined via a *max* function to form a *neuromorphic neuron*. Furthermore, synapses are divided into distal and proximal categories, where proximal synapses act as enable signals, and distal synapses perform a clustering function. At the fourth layer, multiple neurons are combined via winner-take-all inhibition to form a *neuromorphic column*. At the column level [9], an illustrative benchmark concerns a model mouse that learns and navigates a grid containing sparsely placed features. Active dendrites as described here have also been applied to online classification [11] and reinforcement learning [10].

## 3. Neuromorphic Dendrites

What follows is a definition of the *n*D function using the notational language of machine learning. In contrast, the previous above-cited work uses a spiking neuron formulation where information is encoded and processed as bit vectors – a unary system in effect.

### *3.1 Definitions*

Inputs to an *n*D are *feature vectors*. A feature vector, *x*, consists of *m* components (features), each of which has an integer *value* from 1 to *n*. The values for different features do not need to have the same range, but to simplify discussion it is assumed in this document that they are the same.

The function of an *n*D is to cluster input feature vectors according to similarity. To do this, an *n*D is composed of *p templates* that model the dendritic segments. Each template defines a cluster having an associated *cluster identifier* (CId).

Specifically, a template is a vector consisting of *m* components, one for each feature. Each of the components is itself a vector having *n* components, one for each value that a feature can have. The *n* components are interpreted as *weights,* that is, each value that a feature can take is given its own individual weight. The collection of templates is maintained as a three dimensional weight array $w_{1:p\ 1:m\ 1:n}$. Weights range from 0 to $w_{max}$.

As input feature vectors flow through the *n*D, online clustering consists of an inference step, immediately followed by an update (learning) step. During the inference step, input feature vectors are applied to the *n*D and compared with their learned templates. The network output, *z*, is the CId of the template that most closely matches the input pattern in the manner described in the next section.

### *3.2 Inference*

In general, there are two types of features. In one type, the values assigned to a given feature bear no ordered relationship; two features are either the same or not the same: apples = 1; oranges = 2. In the other type, there is an ordering relationship. A feature may be a voltage that ranges from 1 to 32, say. In this case, different values for the same feature can have varying grades of similarity; 27 is similar to 28 and 26, but not 7.

First, consider the apples and oranges case. For input feature vector *x*, the inference step is:

$$z = argmax_{i=1:p} \sum_{j=1:m}(w_{i\ j\ x_j})$$

A detailed example is in Figure 3. To keep the example simple, it contains feature vectors having only three components and, coincidentally, three templates. Each feature value addresses the weight for that value. The value of feature $x_1$ is 3, so in template 1, it accesses the weight for value 3 which is 7. The feature $x_2$ is 1, so it accesses the weight for value 1, which is 3, etc. For each template, the accessed weights are added, and the template with the largest sum is the "winner", and its index is the output, i.e., the CId. In the example, this is template 2. In the event of a tie, the winner can be determined arbitrarily, but in simulations reported here the winner is the template with the smallest CId.

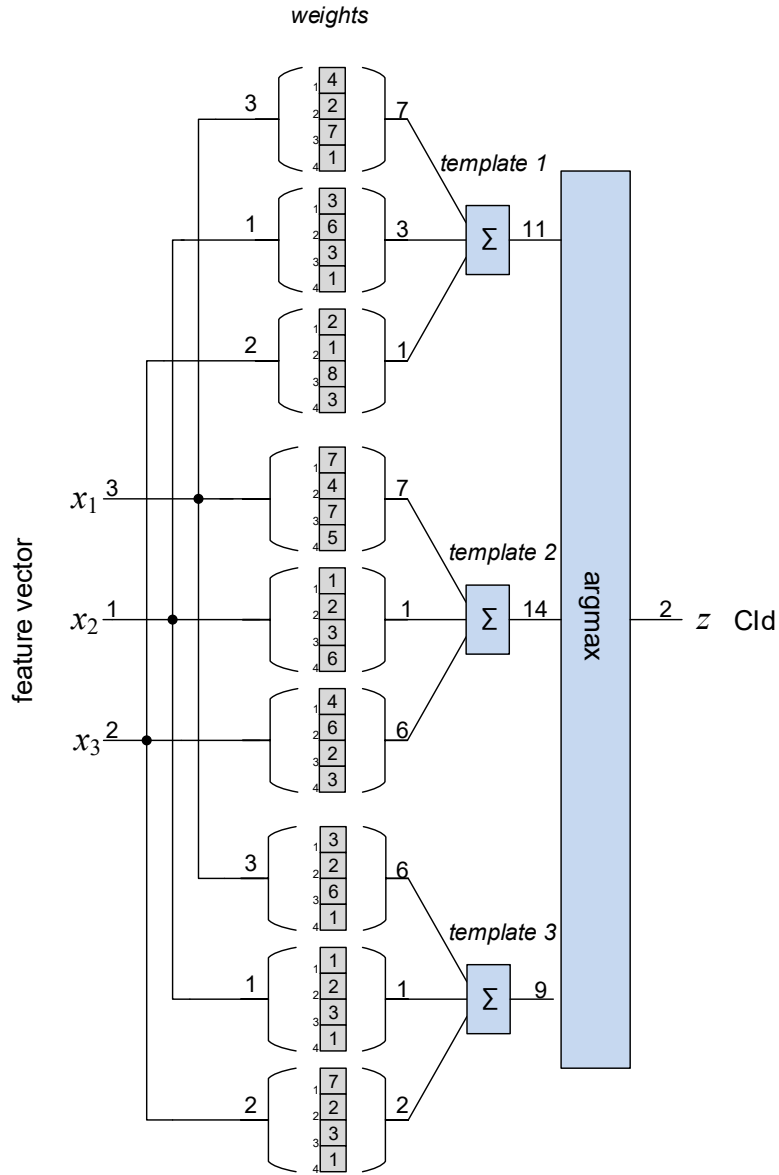

**Figure 3.** *n*D inference example. The input feature vector is [3 1 2 ]. In template 1, the first feature value, 3, indexes the third weight (numbered from the top), thereby accessing weight 7. When all the weights for the templates are summed, template 2 has the largest total weight (14). Consequently, the output is 2.

### *3.3 Update*

A weight *update* occurs immediately after inference. Weight changes depend on the inputs and output of the inference step. The weights associated with the input feature values in the "winning" template *z* are increased by a *capture* amount, and weights of all the other input feature values for template *z* are decreased by a *backoff* amount.

$$w_{z\,j\,x_j} \leftarrow min(w_{z\,j\,x_j} + capture, w_{max})$$

$$w_{z\,j\,k \neq x_j} \leftarrow max(w_{z\,j\,k \neq x_j} - backoff, 0)$$

An example of weight update for the winning template is in Figure 4.

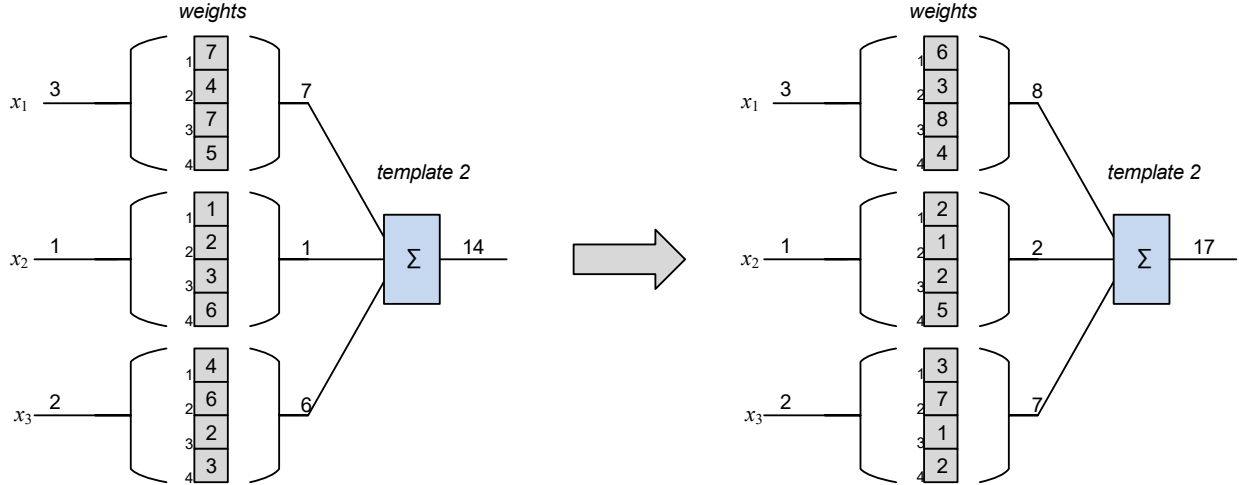

**Figure 4. Winner update before update (left) and after update (right). Weights for input feature values are incremented by the *capture* amount (*capture* = 1 in this example). Weights for all other feature values are decrement by the *backoff* amount (*backoff* = 1 in this example). After update, the input vector [3 1 2] yields a stronger match than before update.**

For adaptivity, a third parameter *search* << *capture* and a base weight $w_{base} < w_{max}$ are introduced. Weights for a non-winning template are incremented by *search* up to $w_{base}$.

$$w_{i \neq z\ j\ x_j} \leftarrow max(w_{i \neq z\ j\ x_j}, min(w_{i \neq z\ j\ x_j} + search, w_{base}))$$

### *3.4 Similarity Coding*

For the situation where feature values are ordered and have degrees of similarity, the value of feature $x_j$ is expanded to include similar features. To be more precise, the *range*, *r*, of an input feature $x_j$ generates an expanded set of values spanning $x_j \pm r$. An example is in Figure 5.

Consequently, for input feature vector *x*, the inference step becomes:

$$z = argmax_{i=1:p} \sum_{j=1:m}(w_{i\ j\ x_j \pm r})$$

The primary update operation is:

$$w_{z\ j\ x_j \pm r} \leftarrow min(w_{z\ j\ x_j \pm r} + capture, w_{max})$$

$$w_{z\ j\ k \neq x_j \pm r} \leftarrow max(w_{z\ j\ k \neq x_j \pm r} - backoff, 0)$$

The search update is:

$$w_{i \neq z\ j\ x_j \pm r} \leftarrow max(w_{i \neq z\ j\ x_j \pm r}, min(w_{i \neq z\ j\ x_j \pm r} + search, w_{base}))$$

Observe that the *n*D uses no multiplications, only additions and subtractions.

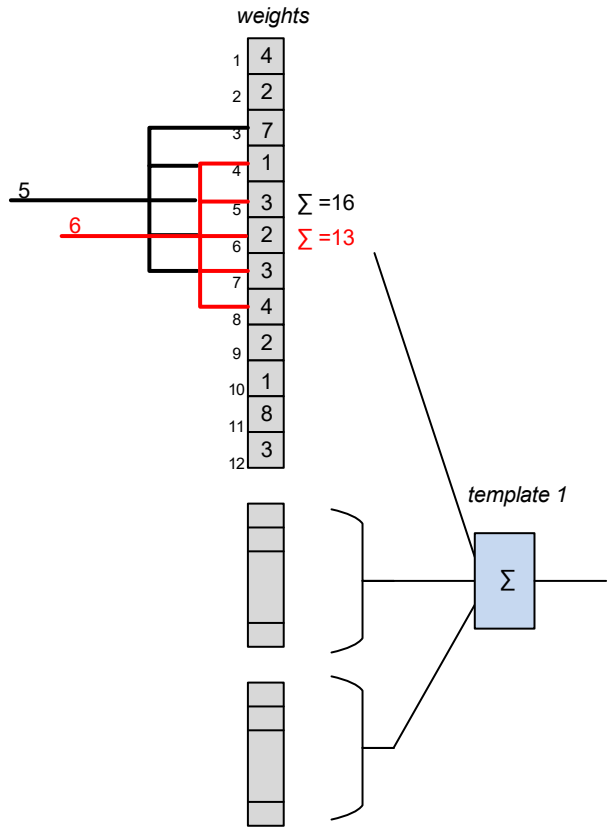

**Figure 5. Example of similarity coding. For input feature value 6 and range = 2 all weights between 4 and 8 are accessed and sum to 13. For input feature value 5, the weights between 3 and 7 sum to 16. The weights that are added overlap in 4 places, so in normal situations, their sums are similar, reflecting feature value similarity.**

### *3.5 Update Intuition*

For a given application, values for *capture*, *backoff*, and *search* are experimentally determined by parameter sweeps. Qualitative relationships among update parameters are discussed here; these can be used to guide parameter sweeps.

First, *backoff* and *capture* control the formation of clusters. In particular, the relative values of *capture* and *backoff* affect the degree of similarity between members of the same cluster. Say the degree of similarity is the minimum number of features that two vectors can have in common in order for them to be assigned to the same cluster. Increasing *backoff* with respect to *capture* tends to increase the degree of similarity required to be a member of the same cluster; decreasing it has the opposite effect.

Applied together, *capture* and *backoff* can cause the weights to converge fairly quickly and deeply. That is, convergence may make the weights fairly inflexible to shifts in feature values as later input vectors are applied. This is an example of the classic stability/plasticity dilemma.

The parameter *search*, in conjunction with *backoff*, provides fluidity to allow templates to adapt their weights as feature values shift. Say the weights for template *T* have been established initially, but later patterns belonging to cluster *T* become sparse or non-existent. In the background, the *search* update gradually increases all the weights belonging to template *T*. Eventually, if a new input vector comes along and the summed weights of none of the other templates exceeds those of *T*, i.e. cluster *T* is the *max*, then template *T* will output its associated CId, and update will begin the process of shifting the template toward a cluster that contains the new vector. On the other hand, if an input vector belonging to template

*T*'s original cluster comes along, then the *backoff* update will push down all the weights not belonging to the applied input, and the original cluster will be reinforced. The weight $w_{base}$ limits how strong a match the search process alone can eventually provide. Hence, a template primed by *search* cannot take over an input vector that is already a relatively strong match with a different template.

### 3.6 Generalization: Multiple Values per Feature

For the remainder of this paper, *x* is a feature vector having *m* components. In the more general case, however, a given feature can be assigned multiple values simultaneously. If the maximum number of simultaneous values is *q*, then *x* becomes an *m* by *q* feature array. Inference is defined as:

$$z = argmax_{i=1:p} \sum_{j=1:m} (\sum_{k=1:q} (w_{ij} x_{jk}))$$

Note that the dimensions of the weight array are the same as before. Updates and similarity coding all extend to this more general case in the obvious way.

### 3.7 Relationship to Spiking Neurons

In an algorithmic implementation of the *n*D as defined above, individual feature values are encoded as binary numbers. However, as an alternative, feature values can be represented in unary form using 1-hot encoding, so the feature vector becomes a bit vector. If the bits are then interpreted as "spikes", the *n*D can be interpreted as a building block for spiking neural networks (SNNs). An SNN approach is taken in prior work by the author [9][10][11].

### 3.8 Relationship to Hawkins's Active Dendrites

In a seminal paper [3] defining active dendrites, Hawkins and Ahmad say: "A permanence value is assigned to each potential synapse and represents the growth of the synapse. Learning occurs by incrementing or decrementing permanence values. The synapse weight is a binary value set to 1 if the permanence is above a threshold."

That is, learning is done via permanence values rather than incrementing and decrementing the weights directly. A synaptic weight is then either 1 or 0, which means that only single bits are added to form a template's output, thereby simplifying the inference function.

## 4. Simulation Framework

For demonstrating its capabilities, *n*D simulations are performed on a spike sorting benchmark.

### *4.1 Spike Sorting*

When an array of electrodes containing multiple sensors is inserted into the brain, or cultured neurons are grown on top of an array of electrodes, there is very little control over the precise locations of electrodes with respect to the locations of neurons. Each electrode is connected to a channel, and when a neuron emits a spike it may be detected on multiple channels. Furthermore, a given channel may detect spikes from multiple neurons. Taken together, spikes from a given neuron can be sorted according to shapes detected at the channels as well as the number and relative locations (patterns) of channel spikes.

In this work, the focus is only on clustering shapes from individual channels. S*hape sorting* provides an excellent demonstration vehicle for an online clustering method. Consequently, the term "shape sorting" will be used for the process of distinguishing neurons according to the shapes of their action potentials as observed on a single channel. Clustering based on multi-channel spike patterns will be discussed in detail in a later document.

### *4.2 Neuron Variability*

What makes the shape sorting problem especially challenging is that the exact physical characteristics and behavior of neurons varies from one to another. Furthermore, the exact spike shapes emitted from a given neuron vary from one spiking instance to another. Figure 6a is the result of experiments with pyramidal neurons [1]. Each individual line is the average spike shape for a particular neuron. The dark line is the average of the averages. These averages illustrate neuron-to-neuron variations. Figure 6b shows how much instance-to-instance variation there can be for a given neuron [13].

Natural variations are studied experimentally by Stratton et al. [12], where they caution the reader that in realistic situations, systematic classification errors can occur. These errors "rather than just obscuring the significance of results, can cause the appearance of results that are entirely fallacious." Because synthetic neurons are studied in this paper, the deleterious effects of natural variations can be studied in a more controlled way, with the caveat that synthetic neurons are only a simplified model for biological neurons.

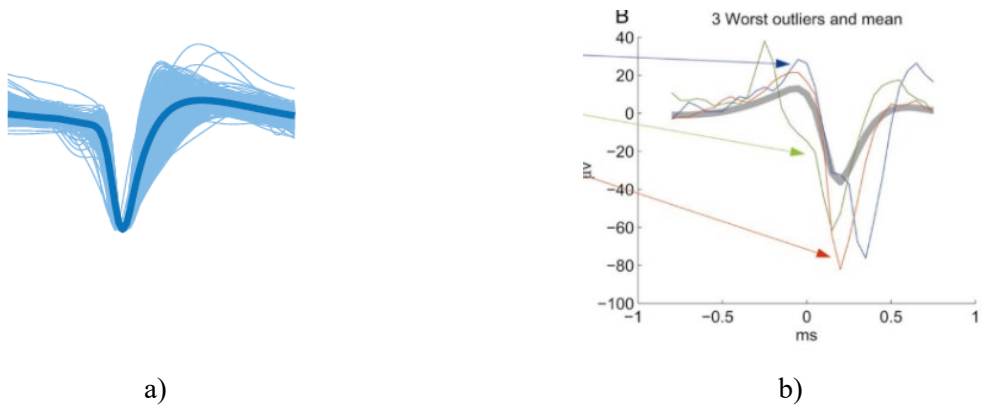

a)                                                                                         b)

**Figure 6. Spike shape variations from extracellular recordings. a) Averages for multiple neurons, from Figure 2C in [13]  b) Instance variations for a single neuron, taken from Figure 3B in [1].**

*Note: Intracellular vs. Extracellular Recording*

Normally, one thinks of a "spike" as positive-going, and this is the case for intracellular recordings of action potentials. See Figure 7 below. However, the waveforms analyzed here are assumed to be from extracellular recordings, which have the opposite polarity: spikes go down.

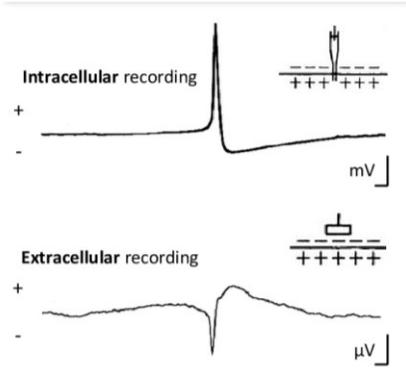

**Figure 7. Intracellular and Extracellular action potential waveforms. Figure excerpted from [5]**

### 4.3 Synthetic Benchmark

A major problem when studying spike sorting using experimentally captured signals from a mass of functioning neural tissue is that there is no practical ground truth for measuring results. Neurons are part of a large, interconnected population, and electrodes must have a certain separation, tens of microns, in order to avoid disturbing or damaging neurons in a way that significantly affects their operation.

Consequently, in this study, synthetically generated neurons and neuron spikes are employed. With this approach, the details of shape sorting can be studied closely in a controlled way, and the merits of the proposed online clustering method can be demonstrated.

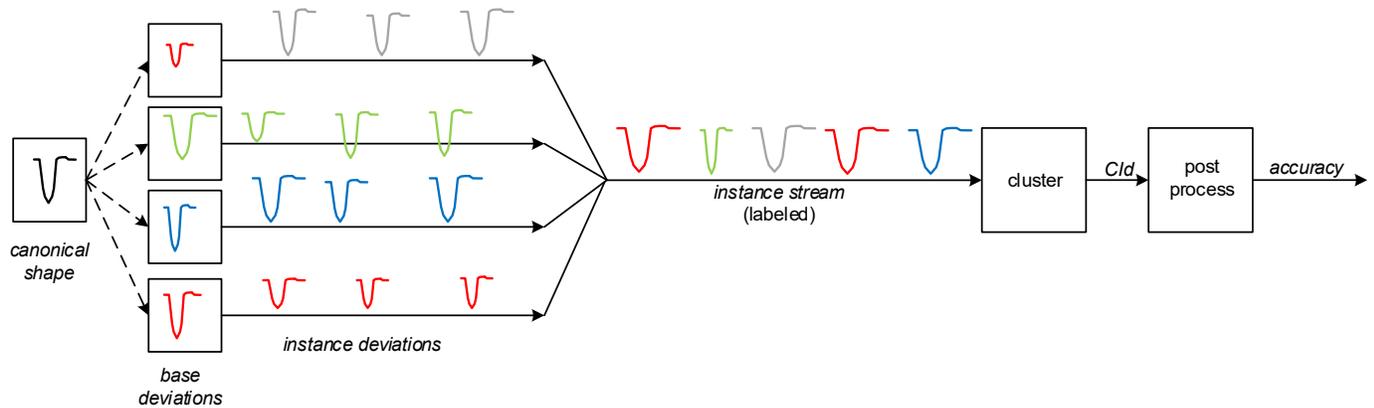

**Figure 8. Simulation framework. A canonical neuron shape is defined. Then base neuron shapes are formed as deviations from the canonical shape. Each cycle, a base neuron is pseudo randomly selected. Assuming features are normally distributed, it generates an instance shape that is a deviation from its base neuron shape. Parameters for base deviation and instance deviation are defined separately. In simulations to follow, input streams typically consist of 10,000 spikes.**

As shown in Figure 8, the simulation process begins with a canonical action potential shape. The base shape is specified as mean values for a set of six features (Figure 9). Given the mean values, base deviations from a normal distribution are used to form the base shapes of subject neurons (four base neurons are shown in Figure 8). Then, to generate a stream of spikes, the base neurons are chosen pseudo randomly, and the chosen neuron produces a spike instance that deviates from the base shape. Overall, this yields a stream of labeled spike shapes, with the label of a shape being the base neuron that produced it. For simulations, both the base deviation and the instance deviations are parameters.

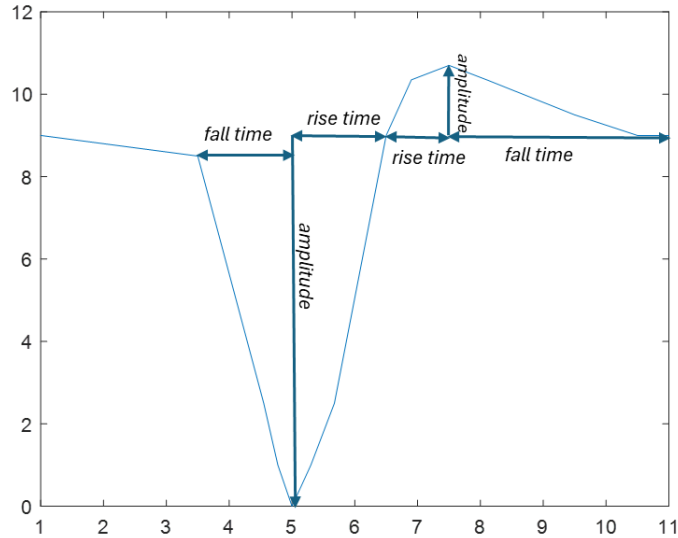

**Figure 9. The canonical spike shape is based on six features that form a feature vector describing the shape. Note that some points in the shape are defined as linear functions of the six features.**

### *4.4 Base Neuron Deviation*

Considering the canonical shape's features as mean values, base neuron shapes are generated for a range of standard deviations from the feature means. It was found that a .375 standard deviation yields a family of waveforms that are similar to experimental waveforms [13]. Refer to Figure 10. The plots for synthetic and experimental waveforms are very similar (and could be made more so, with some additional tweaking). Hence, for simulations to follow the base deviation is set at .375.

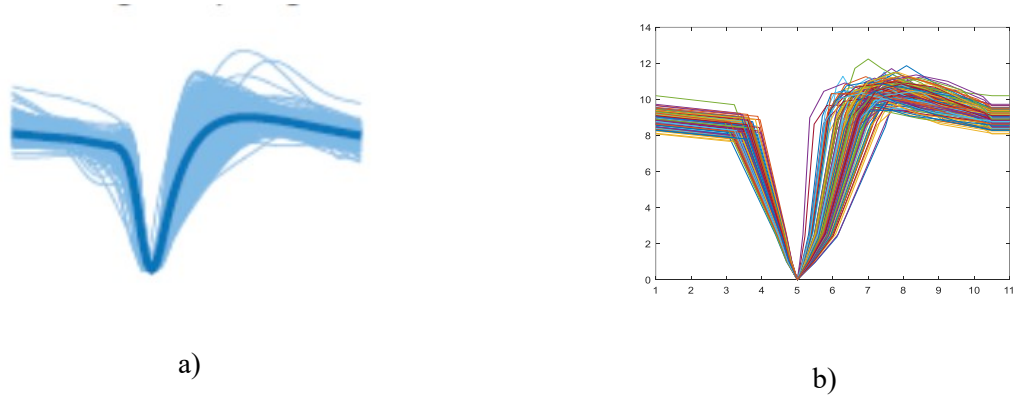

a)

b)

**Figure 10. Variations in base neuron shapes. a) Averages over multiple neurons taken from [13]. b) Shapes for randomly selected 100 base neurons from a normal distribution with .375 standard deviations from canonical feature values.**

*4.5 Accuracy Metric*

The overall objective is to achieve a 1-to-1 map from neurons to CIds, but exactly which neuron to which CId is not important. What is important is how close the overall mapping comes to the ideal 1-to-1. The metric for measuring this is best illustrated via an example (Figure 11). The figure contains a conventional contingency table where each entry is the count of spikes from a given neuron that are mapped to a given cluster, denoted as a CId. Because the mapping must be 1-to-1, a neuron (row) can be used only once, and a cluster identifier (column) can be used only once. Given that, a *sorting accuracy* metric is the *maximum sum of contingency table entries* under the constraint that each row and column can be used only once, *divided by the total number of spikes in the table* (5K for the example shown). This metric can be optimized via the Munkres assignment algorithm[1]. In the figure, the selected entries are highlighted in green. Their sum is 4120, and the sorting accuracy is .82.

|  | Cluster Identifier | | | | | | |
| --- | --- | --- | --- | --- | --- | --- | --- |
|  | CId 1 | CId 2 | CId 3 | CId 4 | CId 5 | CId 6 | assigned |
| Neuron 1 | 1180 | 0 | 1 | 855 | 0 | 0 | 1180 |
| Neuron 2 | 0 | 1012 | 0 | 0 | 0 | 0 | 1012 |
| Neuron 3 | 0 | 0 | 0 | 1 | 0 | 696 | 696 |
| Neuron 4 | 0 | 4 | 539 | 0 | 0 | 0 | 539 |
| Neuron 5 | 0 | 0 | 0 | 0 | 379 | 0 | 379 |
| Neuron 6 | 5 | 0 | 314 | 0 | 0 | 14 | 314 |
|  |  |  |  |  |  | sum | 4120 |
|  |  |  |  |  |  | sort acc. | 0.824 |

**Figure 11. Example table for measuring spike sorting accuracy.**

## 5. *k*-means Baseline

For later comparisons, an offline *k*-means method [4] establishes a performance baseline.

Define a *generated cluster* to be the set of spike shapes generated by a given base neuron. N neurons yield N generated clusters. Because all the pseudo random spike instances have the same deviations with respect to the base neurons, the generated clusters are "spherical" with approximately the same radius. This makes them well-suited for *k*-means clustering. However, due to the aforementioned natural variations there may be large overlaps among the generated clusters. This overlap is a fundamental challenge for any clustering method, including *k*-means. One problem is that with a high degree of overlap, a clustering method may converge to any of a large number of local optima, and these local optima may differ significantly when it comes to cluster quality.

As defined above, a feature vector, *x*, consists of *m* components. For this study, $m = 6$ (Figure 9). Also, for initial simulations, feature values are floating point numbers and floating point arithmetic is used.

Each cluster is associated with a *centroid*: a vector that is a component-wise mean of all the members of the cluster. *k*-means begins with a set of initial centroids and uses these to form a set of clusters, where a feature vector belongs to the cluster with the nearest centroid. Then, it computes the centroid for each of the new set of clusters, and the algorithm repeats with the new set of clusters until convergence is achieved (cluster membership does not change) although here we do not insist on complete convergence. Rather a *convergence metric* is the fraction of vectors whose nearest centroid does not change from one iteration to the next. Convergence = 1 if complete convergence is achieved. In the simulations to follow, the minimum convergence is set at .99.

It is assumed that the number of neurons is known, so the number of centroids is set equal to the number of neurons. The case where the number of centroids does not match the number of neurons is explored later.

---

[1] https://github.com/JieYangBruce/TorqueClustering/blob/main/munkres.m

The first *k*-means simulations initialize the *k* centroids with the same feature values as used for generating the clusters in the first place. This is an ideal case because in reality the feature values of the base neurons are unknown *a priori*. This case is significant because it establishes a practical upper bound on accuracy that reflects the best that can be done given the intrinsic neuron variabilities and resulting cluster overlaps. Simulation results for the ideal case are in Figure 12.

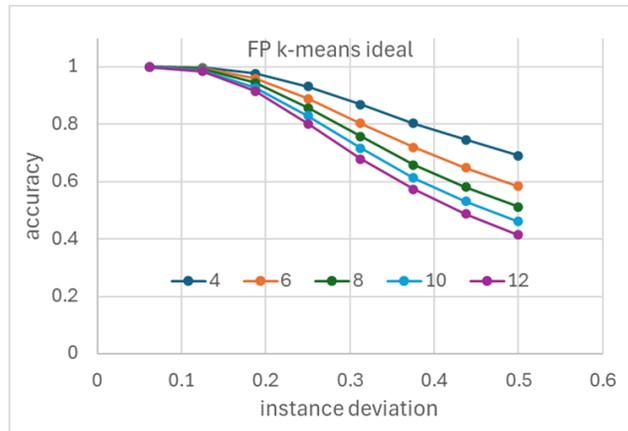

**Figure 12. Simulation results for floating point *k*-means clustering when centroids are initialized to the base neuron feature values. The number of neurons range from 4 to 12. Mean accuracy as a function of instance deviation for neuron counts from 4 to 12. These results place a practical upper bound on accuracy.**

For more realistic *k*-means simulations, centroids are initialized to shapes derived from the canonical shape with standard deviations of .375 from the feature means, but *independent* of the base neuron shapes. In other words, the initial shapes are neuron-like, but different from the base neurons. When this is done, *k*-means converges to a local optimum that is determined by the randomly generated initial centroids, and there can be significant variations in accuracy depending on the random number generator (*rng*) seed. Consequently, for the results reported here, the simulator performs runs with 16 different *rng* seeds, and the accuracies are averaged.

A stream of 10,000 pseudo-randomly generated shapes is divided into 5000 training shapes and 5000 test shapes. The 5000 training shapes are first clustered via *k*-means, then the resulting centroids are used for clustering the 5000 test shapes in a single pass. Configurations with neuron counts between 4 and 12 are simulated. The base deviation is .375 (3/8) as described above, and instance deviations vary from 1/16 to 1/2. The results are in Figure 13.

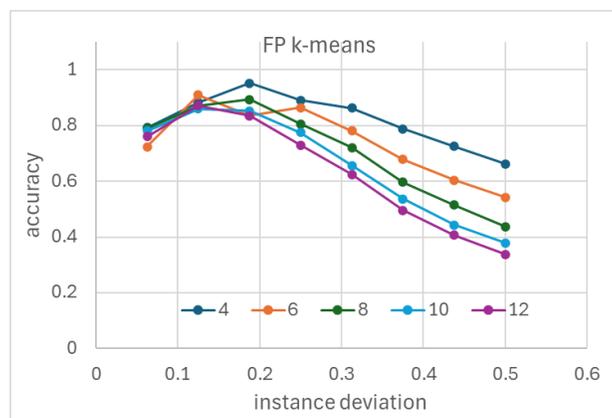

**Figure 13. Simulation results for floating point *k*-means clustering; 16 runs with different *rng* seeds. Simulation conditions are the same as Figure 12.**

The average accuracy is generally better for fewer neurons. This stands to reason because there will be less cluster overlap to contend with. For small instance deviations, average accuracy is between .7 and .9. As the instance variation increases, accuracy increases slightly and then falls off. When the instance deviations are the same as the base deviations, the accuracy is between .5 and .8.

The lower accuracies for the smaller instance deviations are unexpected, but they appear to occur because two (or more) base neurons are often very similar to each other, and when the instance deviations are also small, *k*-means convergence occurs more quickly (i.e. with fewer k-means iterations – see Figure 14 ). This quick convergence to a local optimum allows less opportunity to converge to a better local optimum.

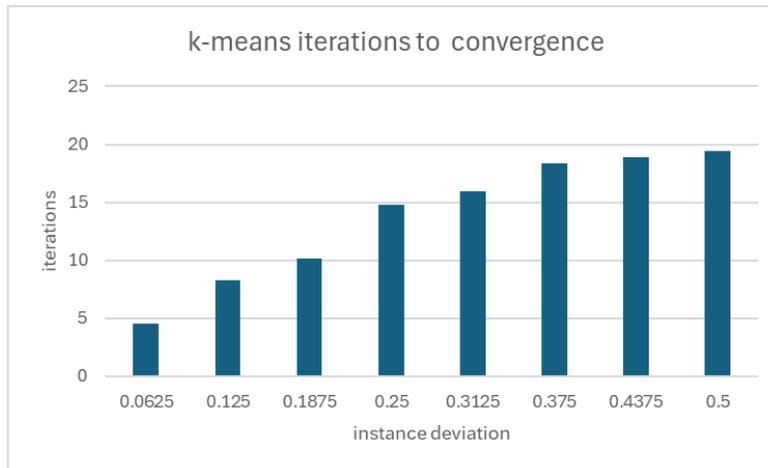

**Figure 14. The average k-means iteration steps to .99 convergence increases as the instance deviation increases. Results are for eight base neurons.**

## 6. Online Clustering

In this section online clustering using the *n*D for shape sorting is studied in detail.

### *6.1 Scaling and Discretization*

In order to make the input suitable for clustering via an integer *n*D, the six features specifying a shape are first normalized so that their maximum ranges are plus and minus three standard deviations from their base feature values. After normalizing, values are scaled and discretized to integers between 1 and 32. After scaling, normalization and discretization, simulation results are in Figure 15. Compared with the original floating point versions (Figure 13) the results are practically the same; very little accuracy is lost in moving to normalized, discretized feature vectors.

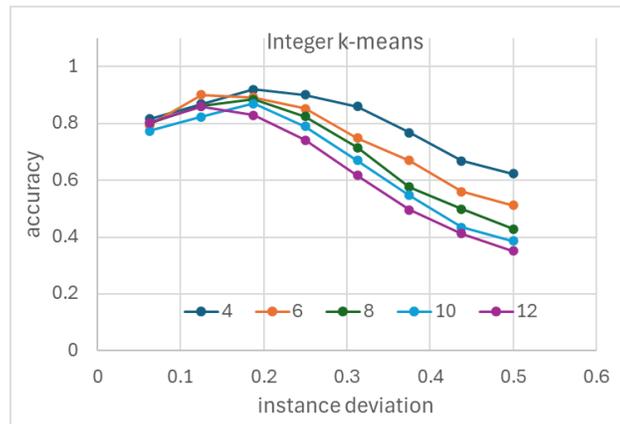

**Figure 15. Simulation results after scaling, normalization, discretization. Simulation conditions are the same as Figure 12.**

### *6.2 Hyperparameters*

There are six hyperparameters (Table 1). In theory, these hyperparameters should be established for each pair of base and instance deviations to reflect a specific application environment. However, in simulations to follow, the base deviations are fixed (at .375) so only instance deviations are varied.

Hyperparameters were determined via simulation sweeps with subjective assessment of the results. After preliminary study, two of the six hyperparameters were fixed for all simulations: the maximum weight $w_{max}$ and the range (*r*). The sweep process consisted of selecting a specific neuron count and instance deviation and optimizing accuracy for the one data point. Because of the high correlations in the accuracy curves, optimal hyperparameters for one data point are also optimal (or nearly so) for many of the others. Therefore, it was found that performing sweeps for a small number of points (as few as one or two) is all that is required. Consequently, for simulations here only two different sets of hyperparameters are used: one for the smallest four instance deviations and the other for the largest four instance deviations. See Table 1. The only difference is in capture and backoff, which directly affect cluster formation (Section 3.5).

Table 1. Hyperparameters for *n*D simulation runs

|  | small instance deviations | large instance deviations |
|---|---|---|
| $w_{max}$ | 32 | 32 |
| $w_{base}$ | 28 | 28 |
| *capture* | 3 | 4 |
| *backoff* | 2 | 1 |
| *search* | 1/16 | 1/16 |
| *radius* (*r*) | 3 | 3 |

### *6.3 nD Simulations*

The same 10,000 spike input stream is applied as for *k*-means. Because the *n*D begins "cold" with an initial set of templates (derived from the same initial centroids as used in the *k*-means simulations), the first 5000 streamed inputs are considered "warmup", so accuracy is measured for the second 5000 inputs as the *n*D continues to operate in an online fashion.

Results are in Figure 16. For small instance deviations, the *n*D method outperforms the *k*-means method significantly. For larger instance deviations the accuracies are virtually the same, so overall the *n*D method performs at least as well as *k*-means and for low instance deviations it performs significantly better.

When compared with the ideal case (Figure 12), the *n*D method gives accuracy very close to ideal for smaller instance deviations, and slightly lower accuracy for larger instance deviations.

Observe that the offline *k*-means method makes between 4 and 20 passes through the 5,000 input training stream. The *n*D method makes a single pass through the combined test and training streams.

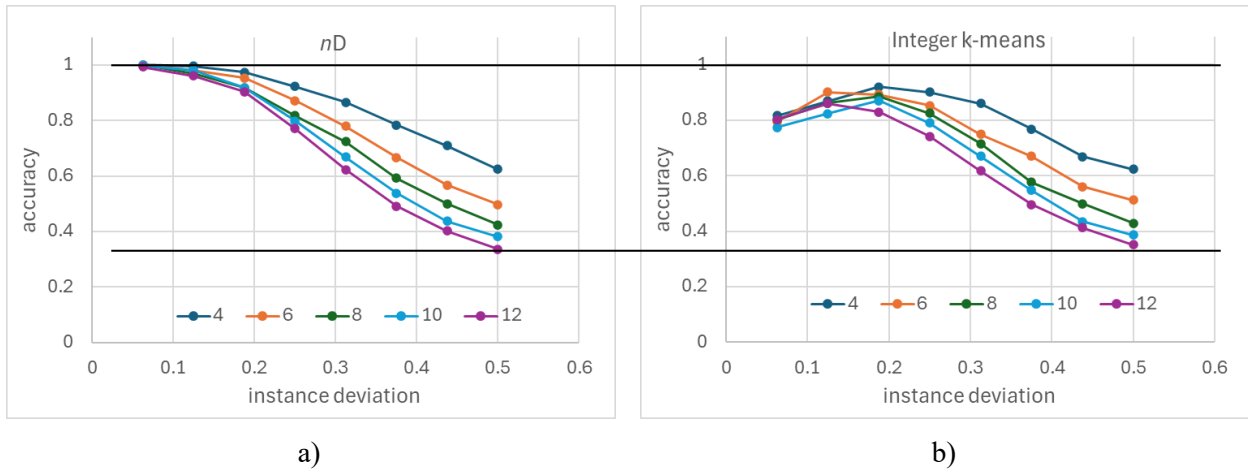

Figure 16. Mean accuracies are for 16 simulation runs with different rng seeds. Simulation conditions are the same as Figure 12. Simulation results for a) *n*D clustering compared with b) *k*-means clustering.

## 6.4 Search

The *n*D results in the previous section were generated with a *search* value of 1/16. If search is turned off by setting it to 0, simulation results are in Figure 17. Accuracies are now similar to the *k*-means accuracies, and in particular, accuracy is reduced significantly for smaller instance deviations. This demonstrates that search provides adaptability by slowing convergence, thereby allowing the *n*D to find a good local optimum.

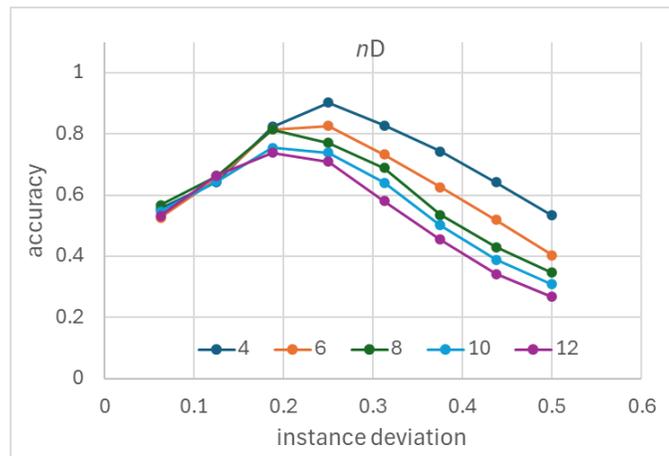

**Figure 17. Mean *n*D accuracy for 16 rng seeds when search is turned off. Neuron and CId counts vary from 4 to 12.**

## 6.5 Probabilistic Search

Because all the other hyperparameters are small integers requiring at most 5 bits of precision, using a search value of 1/16 increases the precision of the weights from 5 bits to 9 bits. Furthermore a search update occurs frequently because it affects a relatively high proportion of the synapses. Hence, search is a relatively expensive operation both in time and space. An alternative implementation uses a search increment of 1, but applies it only 1/16 of the time. To accomplish this an *rng* triggers an actual update for 1/16 of the search cases. Results of simulations using this approach are in Figure 18, compared with search as originally defined. The accuracy curves are virtually identical, indicating the random update approach is a viable way of reducing the number and precision of search updates.

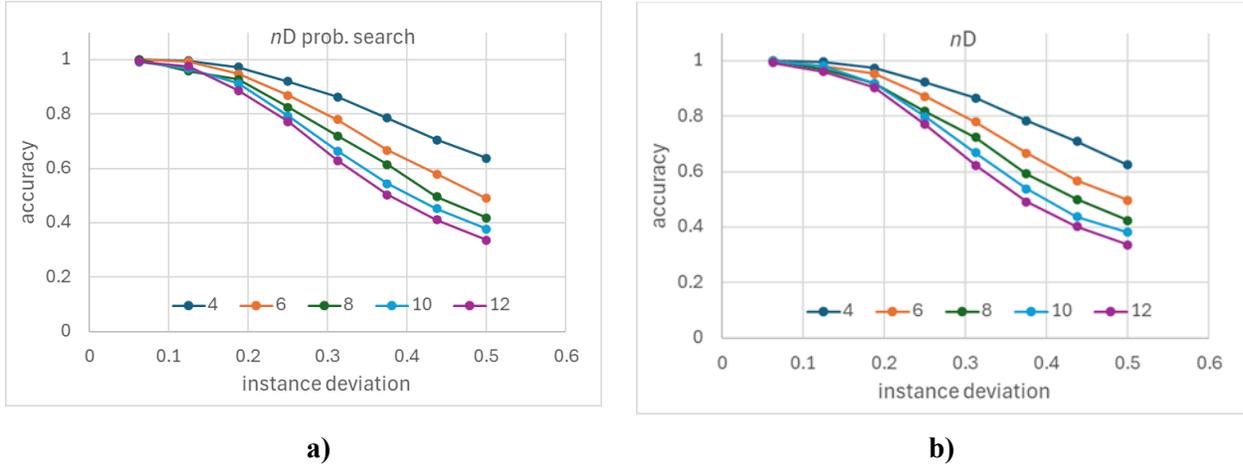

| a) | b) |

**Figure 18. Simulations for neuron counts ranging from 4 to 12 where accuracy is plotted against instance deviations. a) Simulation results with a search of update of 1 applied 1/16 of the time. b) simulation results with a search of update of 1/16 applied every time a search update is called for.**

### *6.6 Energy Consumption (Operation Counts)*

As a proxy for energy consumption, consider the numbers of arithmetic operations (additions) that are required by the four major $n$D sub-functions.

Recall that:
- $m$: number of features
- $n$: number of values that a feature can have
- $p$: number of templates, i.e., the number of clusters
- $r$: range

Assuming similarity coding, the number of operations can be broken into four categories as follows.
- *Inference*: each of $p$ templates adds $m$ values $\pm r$; this requires $p(m(2r+1)-1)$ two-input additions.
- *Capture update*: the winning template performs $m(2r+1)$ additions.
- *Backoff update*: the winning template performs $m(n-(2r+1))$ additions.
- *Search update*: the $p-1$ losing templates perform a total of $(p-1)(m(2r+1))$ additions.

Observe that only backoff is a function of the precision ($\log_2 n$), and only inference and search depend on the number of clusters.

For the simulations performed in this work, $m = 6$, $n = 32$, $r = 3$, $p =$ a variable number of clusters. If we assume $p = 8$, then the total number of operations per feature vector is $328 + 42 + 150 + 294 = 814$ additions. However, these numbers can be reduced significantly by bypassing (skipping) increments of weights that are already at their maximum and decrements of weights that are already at 0. During inference, additions of 0 weights can also be bypassed. If this is done, the total additions per feature vector drops to 481. If probabilistic search is done in conjunction with bypassing additions, the final total is 287 low precision additions per feature vector.

The number of operations for the four categories listed above are plotted in Figure 19. In the end, it is inference that dominates the number of add operations. In contrast to conventional machine learning methods that rely on multi-pass back propagation methods, the $n$D costs of learning are significantly less than the cost of inference. One way of reducing the cost of inference is to use a method akin to the permanence method [3] summarized in Section 3.8.

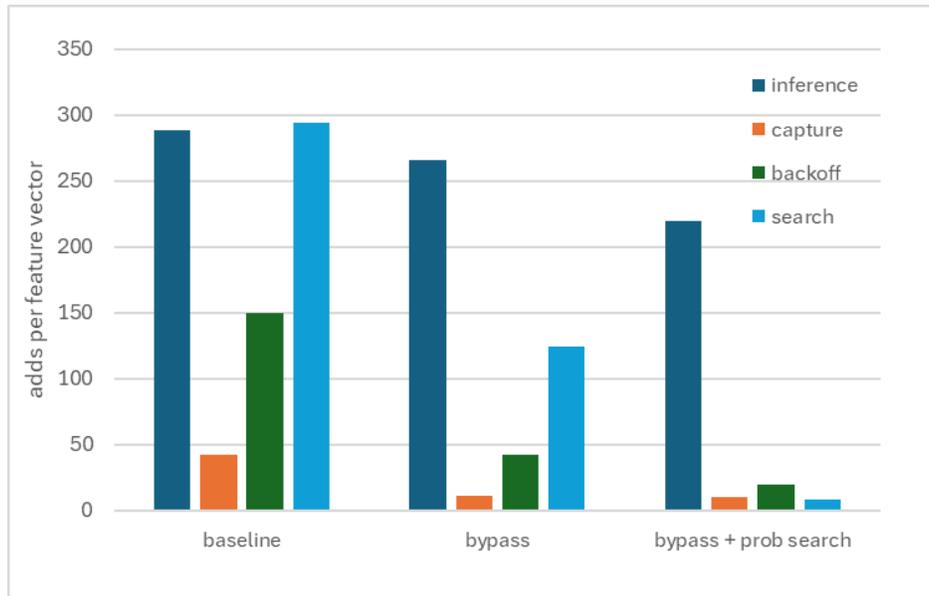

**Figure 19. Add operations per feature vector for the baseline configuration, with bypasses of unnecessary additions, and with bypasses and probabilistic search.**

*6.7 Adaptability*

A key feature of an *n*D is that it dynamically adapts to changes in its input stream. To illustrate this, simulations with 6 neurons were performed where the base neurons were abruptly changed after 5K inputs and then run for another 5K inputs. See Figure 20. Accuracy is computed for every 100 steps. Although for only one example, this behavior is typical. After about 1K inputs, accuracy is near 1. Then when the base neurons are switched so new patterns are observed, accuracy drops precipitously, but then rises and again stabilizes at 1 after about 1K inputs.

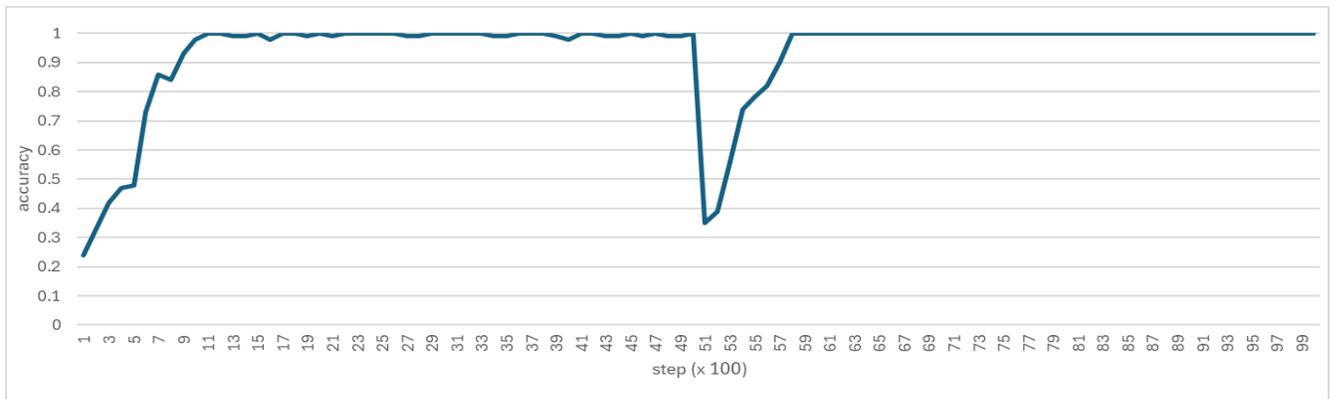

**Figure 20. Accuracy measured every 100 time steps for a simulation of 6 neurons where one set of base neuron features is used initially, and the base neuron features are switched to an independent set after 5K steps.**

## 6.8 Non-uniform Spike Rates

The simulations thus far assume all the neurons spike with equal likelihood, i.e., at the same rate. In many situations, this is not the case. To evaluate performance with different spike rates, a *zipf* distribution of neuron spiking rates is used [7][14].

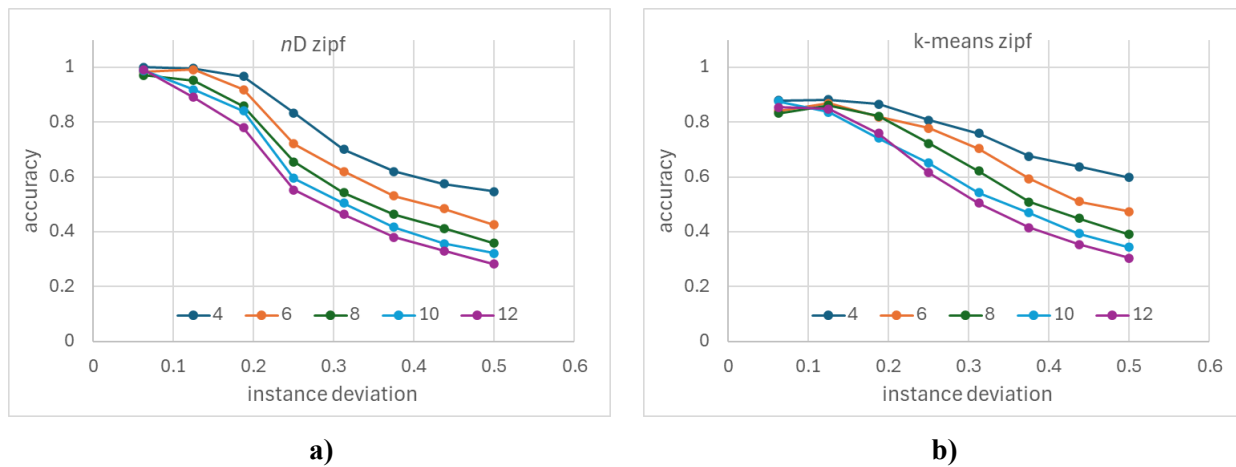

**Figure 21. Simulation results for neurons that spike at rates according to a zipf distribution. Simulation conditions are the same as Figure 12. a) *n*D clustering compared with b) k-means clustering.**

The qualitative conclusions are the same as for equal rates. The *n*D method outperforms *k*-means significantly for small instance deviations and performs on par with *k*-means for large instance deviations.

## 7. Spike Sorting Practical Considerations

The above results apply to the clustering of randomly generated feature vectors. Synthetic spike shapes defined by six features are one easily understood example. Because shapes are synthetically generated, both the number of neurons and labels for evaluating accuracy are readily available. In most real spike sorting applications, however, such information is not available. This section deals (at least partially) with problems posed by this lack of information.

### 7.1 Metric: Number of Accurate Neurons

When multi-channel probes are implanted by experimental neuroscientists, it is unrealistic to require the spiking behavior of all neurons to be observed. Rather, a more practical objective is to maximize the number of neurons that can be *reliably* observed. Consequently, one can define a metric that counts the number of sorted neurons that achieve a minimum acceptable accuracy (*maa*). Accordingly, simulations were performed using the *n*D and zipf firing rates. *maa* values of .8 and .9 are used in the simulation results to follow (Figure 22). Unfortunately, in a real setting the problem is made more difficult because observed spiking behavior does not indicate which of the CIds are associated with accurate neurons.– these would have to be determined by some other means, so results here are essentially an optimistic bound.

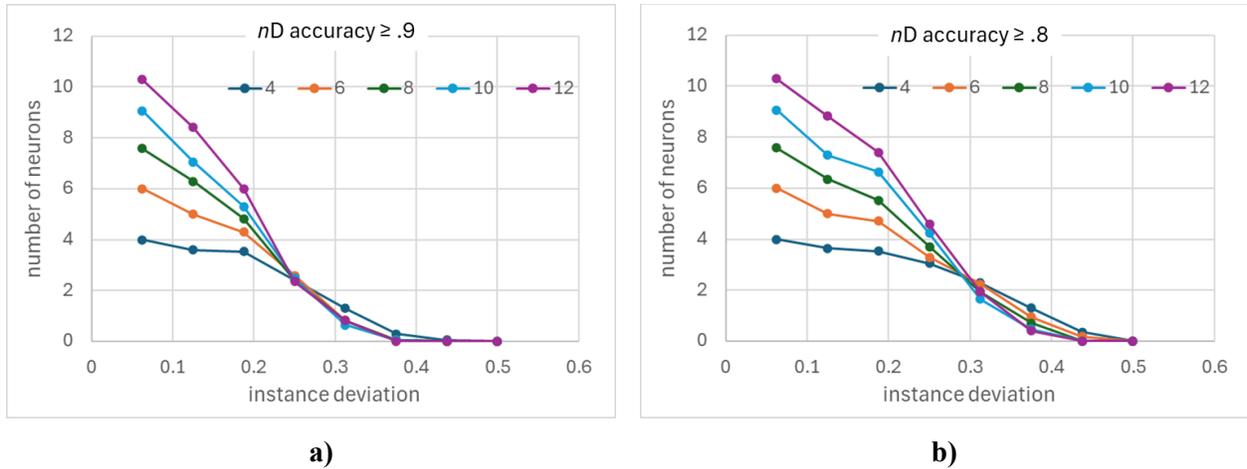

| a) | b) |

**Figure 22. Number of neurons that achieve an accuracy of at least .8 and .9. Simulation conditions are the same as Figure 12.**

Observe that when the instance deviations are very small, most of the neurons achieve good accuracies. However, with deviations starting as low as .1875, the number of accurate neurons falls off dramatically. At a deviation of .25, the number of accurate neurons is the same (4 and 2, respectively) regardless of the number of actual neurons. And it only gets worse for deviations larger than .25. This may be viewed as a negative result; at a minimum it says that the instance deviations must be very small compared with base deviations in order to achieve significant numbers of reliable neurons.

### *7.2 Neuron Count ≠ Cluster Count*

Thus far, the number of simulated neurons and the number of centroids or templates are assumed to be the same. In reality, this is an optimistic assumption because the exact number of neurons is not known in a typical use case.

Consequently, a set of simulations were performed where the number of clusters differs from the number of neurons. Eight neurons are simulated. Results are in Figure 23. For reference, the case where the number of clusters is the same as the number of neurons is shown with the orange line. Interestingly, the case where there are fewer clusters than neurons actually increases the accuracy. Because this is a zipf distribution the two least frequently spiking neurons account for a very small amount of accuracy in the first place, and because they also tend to be less accurate, eliminating them increases the average. For all other cases, the accuracy falls off more-or-less linearly with the number of excess clusters.

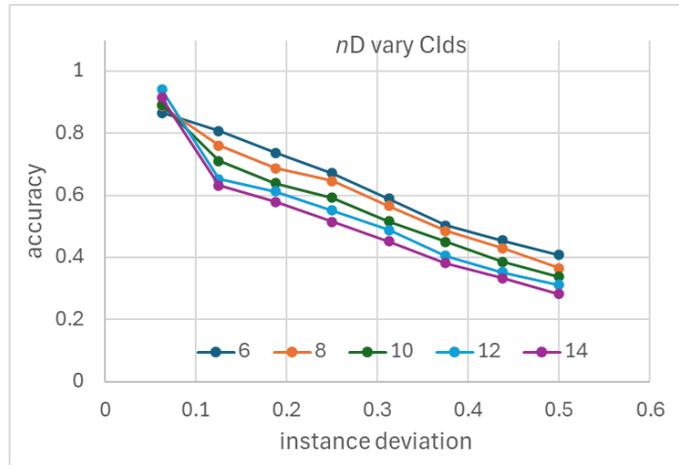

**Figure 23. Simulations with 8 neurons where the number of CIds ranges from 6 to 12.**

These results suggest that an approach for matching the number of clusters with the number of neurons is to begin with a large number of clusters, then merge clusters by ORing the CIds. In this way, multiple CIds may be mapped to the same neuron. A method for determining which CIds should be merged is not considered here; rather, an oracle is used. Hence, consider a metric that measures *potential accuracy* if clusters can be optimally merged. The resulting metric is essentially the clustering purity (Figure 24). In the figure, clusters 1 and 4 can be combined, and the purity (potential accuracy) improves significantly to .93, compared with the accuracy of .82 from Figure 11.

|  | Cluster Identifier | | | | | | |
|---|---|---|---|---|---|---|---|
|  | Cld 1 | Cld 2 | Cld 3 | Cld 4 | Cld 5 | Cld 6 | assigned |
| Neuron 1 | 1180 | 0 | 1 | 855 | 0 | 0 | 2035 |
| Neuron 2 | 0 | 1012 | 0 | 0 | 0 | 0 | 1012 |
| Neuron 3 | 0 | 0 | 0 | 1 | 0 | 696 | 696 |
| Neuron 4 | 0 | 4 | 539 | 0 | 0 | 0 | 539 |
| Neuron 5 | 0 | 0 | 0 | 0 | 379 | 0 | 379 |
| Neuron 6 | 5 | 0 | 314 | 0 | 0 | 14 | 0 |
|  |  |  |  |  |  | sum | 4661 |
|  |  |  |  |  |  | purity | 0.9322 |

**Figure 24. Example contingency table for clustering purity.**

Simulation results with optimal CId merging are in Figure 25. When the number of CIds exceeds the number of neurons, the potential accuracy is nearly the same, regardless of the number of excess CIds. This suggests that overprovisioning the number of CIds does not, in itself, affect accuracy, *providing* optimal merging can be done. However, the problem of determining which CIds to merge remains.

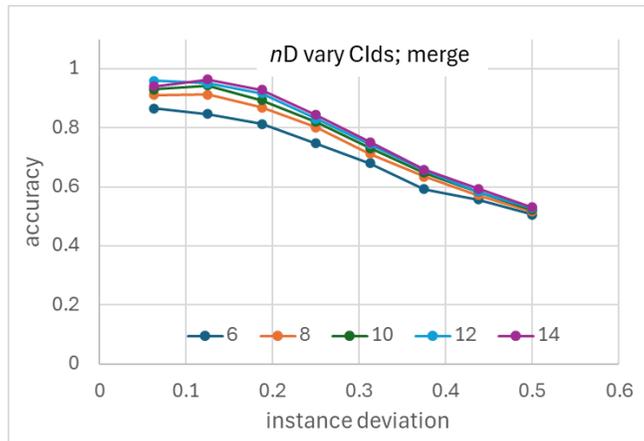

**Figure 25** Simulations with 8 neurons where a variable number of CIds are optimally merged.

## 8. Conclusions

An *n*D out-performs *k*-means over a wide range of neuron counts, instance variations, and spike rates. Furthermore, the *n*D method is online and makes only one pass through the input stream. *k*-means is offline and requires multiple passes (between 4 and 20) for the simulations reported here.

The importance of the search feature of the *n*D method is clear. For small instance variations, the accuracy is much higher with search turned on as compared with search being turned off. To reduce costs, a probabilistic search method is effective. In addition, observe that search does not have to be enabled all of the time. Depending on the application, it can be turned on for initial learning and then switched off-and-on intermittently to update weights when there is the possibility of a change in the input streams. (For that matter, all updates can be managed this way).

With regard to neuromorphic architecture, clustering is done with a *single dendrite*, i.e., at the sub-neuron level. This highlights the potential computing power of a single neuron when composed of multiple active dendrites, as compared with the classical point neuron model. This indirectly supports biological plausibility because, as suggested by Hawkins and Ahmad [3], it answers the question: why do neurons have thousands of synapses?

With regard to the shape sorting application, it is important to note that synthetic neurons are used, so conclusions can only go so far, but they at least give a qualitative idea for the phenomena that are taking place – an upper bound on accuracy caused by natural variations, for example. In some cases the upper bound on accuracy may be so low that shape sorting provides little additional value versus using only spiking patterns across multiple channels, especially if one considers the number of accurate neurons (Figure 22). Hence, it appears important that one should establish the extent of natural variations that are present in a given experimental environment before relying on shapes for accurate spike sorting.

With regard to hardware costs, a detailed analysis has not been done. However, it is worth pointing out that this is a low precision integer model. For the simulations reported here, only five bits of precision need to be maintained in the weights and associated update hardware (assuming probabilistic search). The inference hardware requires more than five bits, but the precision of the adder is still likely to be small (less than 10 bits, say) for practical applications.